\documentclass[runningheads]{llncs}

\usepackage{amsmath}
\usepackage{amsfonts}
\usepackage{graphicx}
\usepackage[square,numbers]{natbib}
\usepackage{amssymb}
\usepackage{algorithm}
\usepackage{algorithmic}
\usepackage{array}
\usepackage{booktabs}
 \usepackage{bm}
 \usepackage{xcolor}

\begin{document}
\title{Generating Math Word Problems from Equations with Topic Consistency Maintaining and Commonsense Enforcement}
%
%
\author{Tianyang Cao\inst{1,2}\and
Shuang Zeng\inst{1,2}\and
Songge Zhao\inst{1}\and Mairgup Mansur\inst{3} \and Baobao Chang\inst{1,2}}
\authorrunning{Tianyang Cao et al.}
%
\institute{ Key Laboratory of Computational Linguistics, Peking University, MOE, China \and
 School of Software and Microelectronics, Peking University, China \and
Sogou Technology Inc.\\
\email{\{ctymy,zengs,zhaosongge,chbb\}@pku.edu.cn,maerhufu@sogou-inc.com}}
\maketitle              
\begin{abstract}
Data-to-text generation task aims at generating text from structured data. In this work, we focus on a relatively new and challenging equation-to-text generation task – generating math word problems from equations and propose a novel equation-to-problem text generation model. Our model first utilizes a template-aware equation encoder and a Variational AutoEncoder (VAE) model to bridge the gap between abstract math tokens and text. We then introduce a topic selector and a topic controller to prevent topic drifting problems. To avoid the commonsense violation issues, we design a pre-training stage together with a commonsense enforcement mechanism. We construct a dataset to evaluate our model through both automatic metrics and human evaluation. Experiments show that our model significantly outperforms baseline models. Further analysis shows our model is effective in tackling topic drifting and commonsense violation problems.

\end{abstract}

\section{Introduction}

Text generation, aiming to automatically generate fluent, readable and faithful natural language text from different types of input, has become an increasingly popular topic in NLP community.

Many recent text generation approaches ~\cite{gyawali2014surface,wiseman2017challenges,puduppully2019data-to-text,chen-etal-2019-enhancing,gong-etal-2019-enhanced,zhao-etal-2020-bridging} focus on data-to-text generation task, which generates textual output from structured data such as tables of records or knowledge graphs (KGs).
However in this paper, we focus on a relatively new type of data-to-text generation task: generating math word problems (MWP) from equations ~\cite{zhou2019towards}, which seems has not been fully studied by NLP community.
Successful math word problems generation has the potential to automate the writing of mathematics questions. 
Thus it can alleviate the burden of school teachers and further help improve the teaching efficiency.
\begin{figure}[htbp]
\centering
\begin{minipage}[t]{0.48\textwidth}
\centering
\includegraphics[width=1.0\linewidth]{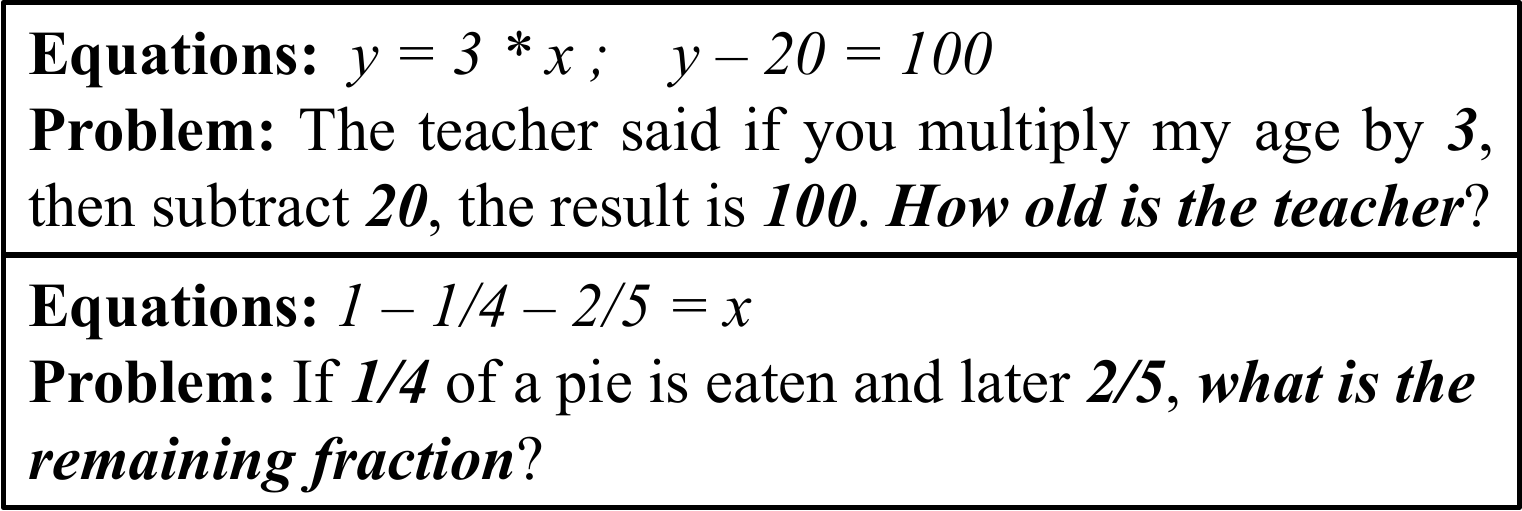}
 \caption{Two examples selected from the MWP generation dataset.}
 \label{table:runing-examples}
 \end{minipage}
 \begin{minipage}[t]{0.48\textwidth}
\centering
 \includegraphics[width=1.0\linewidth]{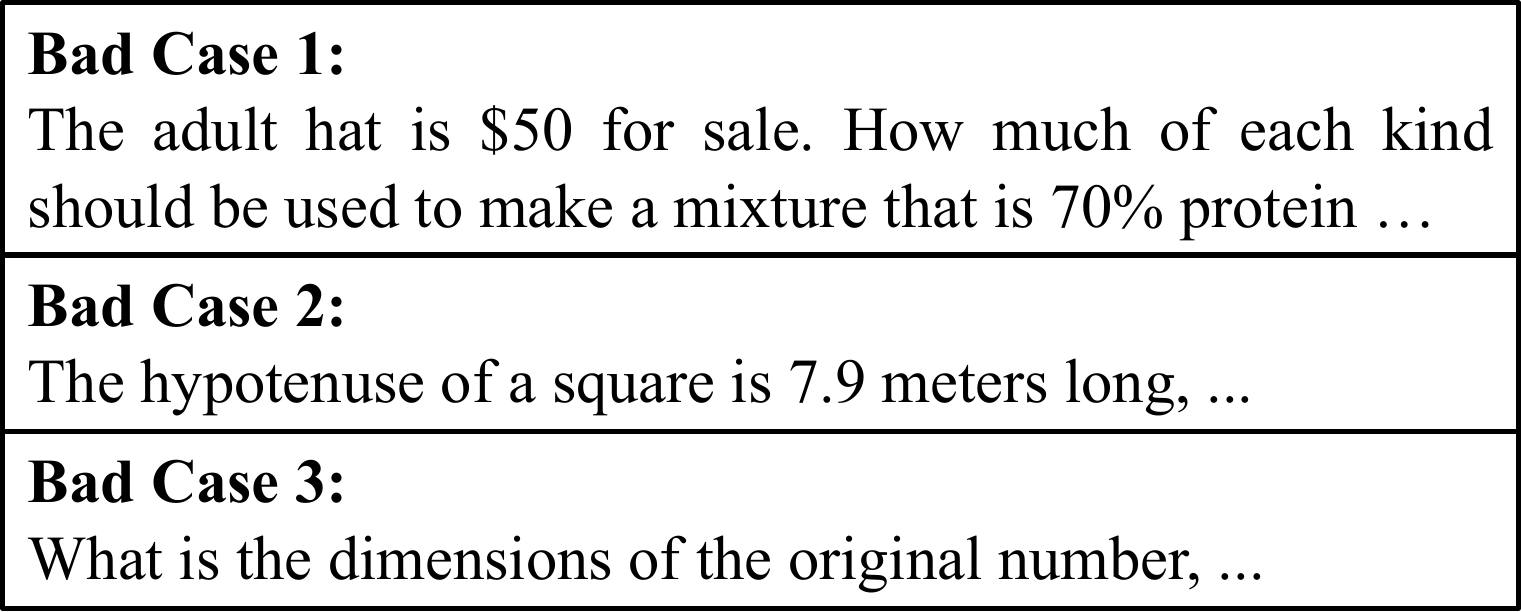}
 \caption{Three bad cases generated by baseline (Seq2Seq) model.}
 \label{table:bad-cases}
 \end{minipage}
 \end{figure}
 
 Our target is to design a model to generate math problem text from the given equations and the generated math problem could be solved by the equations.
Different from the traditional text generation task, there are three major challenges in effective MWP generation from equations: \\
\noindent
(1) Encoding math equations is much more difficult than encoding plain text, tables or KGs. 
A math equation consists of different type of tokens, such as number, variable and operator. 
They express different meanings and are very abstract for generating text. So a model should use different methods to encode them and need to bridge the gap between the abstract math tokens and natural language text.
\noindent
(2) Recent language modeling advancements indeed make generated text more fluent, but still lacking of coherence, especially in the aspect of topic drifting, has always been a non-trivial problem that traditional text generation models usually suffer from ~\cite{cui2020mutual}. 
And we find this problem is even worse in MWP generation since target math word problems in MWP dataset covers a broad variety of topics. 
Figure~\ref{table:bad-cases} shows three bad cases generated by a Seq2Seq model.
The first case reveals the topic drifting problem where the topic of the first generated sentence is the \textit{price of goods} but the topic of the second sentence is changed to \textit{substance mixture}. So how to maintain the topic consistency in generated text to avoid topic drifting is very challenging. 
\noindent
(3) The task requires generated problem text to be in line with the commonsense which is very hard for existing architecture. As shown in the last two cases in Figure~\ref{table:bad-cases}, we cannot say ``hypotenuse of a square" or ``dimension of a number", because they aren't in accordance with the commonsense. So we should design an effective architecture to avoid commonsense violation problem.

\par To tackle these challenges, we propose a novel architecture for generating MWP from equations. First, to effectively encode different kinds of math tokens in the given equations, we propose a template-aware equation encoder that considers both template information and equation information. We further utilize a problem-aware {V}ariational {A}uto{E}ncoder (VAE) with a Kullback-Leibler distance loss to bridge the gap between abstract math tokens and problem text. Then we propose a topic selection mechanism that selects a fixed topic for each generated text and a topic controlling mechanism that controls the topic at every decoding step to avoid topic drifting problem. To cope with the possible commonsense violation 
issue of generated text, we design a pre-training stage as well as a commonsense enforcement module to encourage our model to generate math problem text that is in line with the commonsense. 
 
 \par Our contributions can be summarized as follows:
\begin{itemize}
\item We propose an effective way of encoding different math tokens and a problem-aware VAE to bridge the gap between abstract math tokens and generated text.

\item We utilize a topic selection and a topic controlling mechanism, so topic consistency of generated math problem text could be maintained.
\item We design a pre-training stage and a commonsense enforcement module to alleviate commonsense violation.
\end{itemize}
\par In order to verify the effectiveness of our model, we construct a dataset by obtaining math word problems and their corresponding equations from \textit{Yahoo!}\footnote{https://github.com/caotianyang/math2textcs1}. Experimental results on this dataset show our model significantly outperforms previous models. Further analysis and human evaluation show our model is effective in tackling the three challenges mentioned before, especially the topic drifting and commonsense violation problems.

\section{Task Definition}
\par The input of MWP generation task is a set of equations $\left\{E_1 , E_2, ..., E_{|E|}\right\}$, each equation can be denoted as a sequence of math tokens: $E_k=x_1x_2...x_{|E_k|}$, where $|E_k|$ is the length of $k$-th equation measured by the number of math tokens. Each math token belongs to one of the following three types: math operator (e.g., $+, -, *, ÷, =, ...$), number (e.g., $0.2, 1, 30, ...$), variable (e.g., $x, y, z, ...$). The output of the task is the MWP text: $\bm{y}=y_1y_2...y_L$, which could be solved by the input equations. $L$ is the the length of problem text. Our model aims to estimate the following conditional probability depending on equations and previously generated words $\bm{y}_{<t}$:
\begin{align}\label{eq1}
P(\bm{y}|\bm{x}) &= \prod_{t=1}^L P(y_t|\bm{y}_{<t},E_1 ,E_2,...) 
\end{align}

The difficulty of the input equations in this task is not beyond middle school level, only involving algebra operation in elementary mathematics: ``$+, -, *, /, \wedge, ...$".

\section{Model}
\par
\begin{figure*}
\centerline{\includegraphics[width=1.0\linewidth]{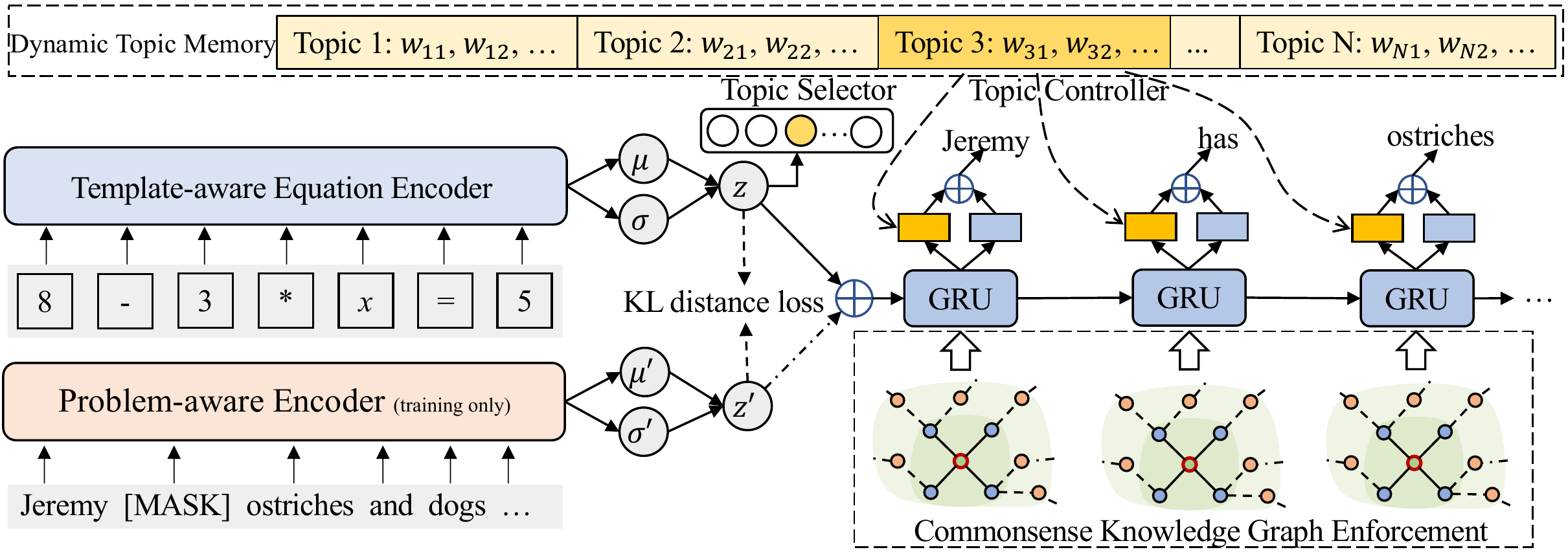}}
\caption{The overview of our proposed model. We omit the pretraining step for simplicity. In our variational autoencoder enhanced model, the problem encoder serves as prior network and the equation encoder serves as posterior network. Topic type predicted by the hidden equation representation $z$ is used to select the corresponding row in topic memory. Next, the MWP decoder resorts to both the dynamic topic memory and the Commonsense KG reasoning to generate MWP text.}
\label{figure:model}
\end{figure*}

 The overall architecture of our model is shown in Figure~\ref{figure:model}. We start with a variational encoder-decoder model as our base model which consists of a template-aware equation encoder and a math word problem generation decoder. Since the math tokens in the original input equation are very abstract and lack enough context information for generating text, we introduce a problem-aware Variational AutoEncoder to encourage the equation encoder to 
produce text-sensitive representation that is more suitable for decoding problem text.


To tackle the problem of topic drifting, we introduce a topic selector with a topic controller. The topic selector chooses a specific topic based on the latent representation of equations. The dynamic topic memory is used to control the decoding process to favor the topic-consistent text. To alleviate the commonsense violation, we introduce a pre-training step to produce commonsense embedding for words and use a commonsense enforcement module to aggregate commonsense information which will influence the following choice at each step of decoding. 
\vspace{-1.2cm}

\subsection{Variational Encoder-Decoder Module\label{subsec:Variational-Encoder-Decoder-Module}}

\par As we mentioned before, we choose the variational encoder-decoder model as the basic model to generate the MWP text from equations. The backbone of our model consists of a template-aware equation encoder and a problem text decoder. \\
\textbf{Template-aware Equation Encoder}: The input to our model is a sequence of math tokens $x_1x_2...x_n$, our input encoder encodes each token to one fixed hidden vector. Math equations encoding is different from encoding other natural languages, we should distinguish numbers, variables and operations to assign them different encoding. 
\par We exploit BiGRU as the basic module, it consumes token embedding of the equation sequences and the hidden states are computed by: $\overleftarrow{\bm{h}}_i = GRU(emb(x_i),\overleftarrow{\bm {h}}_{i-1})$, $\overrightarrow{\bm{h}}_i = GRU(emb(x_i),\overrightarrow{\bm {h}}_{i-1})$. $emb(x_i) = \bm{E}_{token}(x_i)+\bm{E}_{type}(x_i)$ is the sum of corresponding token embedding and type embedding. Combining forward and backward state yields $\bm{h}_i = \overleftarrow{\bm{h}}_i+\overrightarrow{\bm{h}}_i$. 

To improve the generalization capacity of the equation encoder, we further incorporate a soft gate controlled by equation template. The equation template is constructed by replacing all numbers in the equation to a fixed mask [M]. We separately feed the original sequence and the template sequence into two different GRUs, then encoded hidden states are denoted as $\left\{\bm{h}_{a,1},\bm{h}_{a,2},...,\bm{h}_{a,n}\right\}$ and 
$\left\{\bm{h}_{b,1},\bm{h}_{b,2},...,\bm{h}_{b,n}\right\}$, respectively. Here we utilize Gated Linear Unit (GLU) ~\cite{dauphin2016language} to compute final encoded hidden state:
\begin{align}
\bm{h}_k = MLP_1(\bm{h}_{a,k})\odot \sigma(MLP_2(\bm{h}_{b,k})) \ 1\le k \le n
\end{align}
where $\sigma(\cdot)$ is a sigmoid function and $MLP(\cdot)$ is a linear layer. $\odot$ indicates element-wise multiplication. $\bm{h}_{b,k}$ can be understood as a weight matrix to select salient information in $\bm{h}_{a,k}$. We perform linear transformation to $\bm{h}_n$ and approximate mean and variance of $z$'s posterior distribution by assuming the hidden equation representation $z$ follows multivariate Gaussian distribution.
\begin{align}\label{eq6}
\left[\bm{\mu},\bm{\sigma}\right]=MLP(\bm{h}_n) \ \ z|\bm{x}\sim \mathcal{N}(\bm{\mu},\bm{\sigma}^2 \textbf{I})
\end{align}
$\textbf{I}$ is an identity matrix and $z$ can then be sampled by using reparameterization
trick: $\bm{z}=\bm{\mu}+\bm{r}\odot \bm{\sigma}$, where $\bm{r}$ is a standard Gaussian distribution variable.\\
\textbf{Problem Decoder}: For generating 	problem text, we use GRU based decoder. We first initialize the decoder state by $\bm{s}_0=MLP(\left[\bm{h}_n;\bm{z};\bm{h}_n\odot \bm{z}\right])$. We denote the hidden state of the decoder at $t$th step as $\bm{s}_t$ and context vector obtained by attentions over the input equation as $\bm{c}_t$. Assume the decoder generates word $w_{t-1}$ in step $t-1$, 
decoding process can be formulated by :

\begin{align}
\label{eqd}\bm{s}'_t = f(\bm{s}_t) \ \ \bm{s}_t &= GRU(\bm{s}_{t-1},g(\bm{e}_{w_{t-1}})) \\
p_D(y_t|y_{<t},\bm{x},z,\hat{p};\theta_D)&= softmax(\bm{W}^o tanh(\bm{W}^{vs}\left[\bm{s}'_t;\bm{c}_t\right]))
\end{align}
where $\bm{W}^{vs}\in \mathbb{R}^{d \times d}$, $\bm{W}^o \in \mathbb{R}^{d\times |V|}$. $|V|$, $\hat{p}$ and $d$ is the vocabulary size, topic category and embedding size, respectively. $f(\cdot)$ and $g(\cdot)$ is designed for leveraging topic restriction and commonsense restriction, respectively, which will be explained later. We further
adopt copy mechanism ~\cite{see2017get} to copy numbers from equations.
\vspace{-0.5cm}
\subsection{Enhancing Equation Encoder by Variational Autoencoder\label{subsec:enhancing}}
\par Hidden equation representation $\bm{z}$ derived by \eqref{eq6} fails to capture interaction between equations and MWP text. We thus introduce a problem-aware VAE to further restrict $\bm{z}$ into similar vector space of MWP text to obtain problem text aware representation. In this paper, the VAE is comprised of the problem encoder and the problem decoder. As the problem
text is known when training, posterior distribution of $z$ generated by the equation encoder is conditioned on prior distribution generated by the problem encoder. 
\par The problem encoder summarizes the MWP text to a vector $\bm{q}$ and works as a prior network. It takes the corrupted version of problem text $\bm{y}$ as input to guarantee robustness when testing, i.e., we randomly mask and delete some words in the original MWP text. We implement the problem encoder module based on convolutional neural network (CNN) with $F$ different convolutional kernels to extract multi-scale features:

\begin{align}
\bm{h}^q_k &= MaxPool(f_{conv}(\left[\bm{y}_i;\bm{y}_{i+1};...;\bm{y}_{i+l_k-1}\right])) \\
\bm{q} &= tanh(\bm{W}^q\left[\bm{h}^q_1;\bm{h}^q_2;...;\bm{h}^q_F\right])
\end{align}

\noindent where $\bm{W}^k \in \mathbb{R}^{dl_k}$ is the $k$th convolutional kernel and parameter matrix $\bm{W}^q \in \mathbb{R}^{Fd\times d}$. Similar to \eqref{eq6}, we perform linear transformation to $\bm{q}$ and obtain mean and variance of $z$'s prior distribution: $\left[\bm{\mu}',\bm{\sigma}'\right]=MLP(\bm{q}) \ \ z'|\bm{y}\sim \mathcal{N}(\bm{\mu}',\bm{\sigma}'^2 \textbf{I})$.

We denote the problem decoder parameterized by $\theta_D$ as $p_D(\bm{y}|\bm{x},z,\hat{p};\theta_{D})$,
during training, $\bm{z}$ is obtained by prior network. We aim to minimize Kullback-Leibler distance (KL loss) between prior distribution and posterior distribution. Loss function of our Variational Encoder-Decoder framework can then be computed by combining KL loss and generator decoding loss:
\begin{align}\label{eq10}
\mathcal{L}_{VAE}&=-KL(p(z|\bm{y})|| p(z|\bm{x})) \nonumber \\
&+ \mathbb{E}_{z \sim \mathcal{N}(\bm{\mu}',\bm{\sigma}'^2 \textbf{I})} p_D(\bm{y}|\bm{x},z,\hat{p};\theta_{D})
\end{align}
Besides, we use KL cost annealing to avoid KL-vanishing phenomenon ~\cite{bowman2016generating}. During inference, $\bm{z}$ is approximated by posterior network. 

\subsection{Topic Selection and Controling}
\par Generally speaking, given an input equation, for example, $0.5*x+0.3*y=10$, our model should first select a certain type of topic and then incorporate related topic words under this type into the problem decoder.\\
\textbf{Topic Selection}: To leverage topic background to the hidden equation representation $\bm{z}$, we apply an unsupervised document topic model-- Latent Dirichlet Allocation (LDA) ~\cite{blei2003latent} to assign a topic type for each math problem text. We treat each math question as a document, each document is associated with a topic distribution over all topics, meanwhile each topic contains several words with the highest probability in this topic. 
We then estimate the problem topic type through $\bm{z}$:
\begin{align}\label{eq9}
\hat{p}=\arg\max softmax(\bm{W}_z\bm{z}+\bm{b}_z)
\end{align}
\textbf{Topic Controling}: Topic controling renders our generator to interact with topic word distribution. With the help of LDA, a topic memory $\bm{C}\in \mathbb{R}^{|P|\times K\times d}$ is constructed for storing pretrained embedding of topic keywords, where $|P|$ is the total topic number. K means each row of $\bm{C}$ contains information of top-K words of one topic and $d$ is the vector dimension. With the most probably topic type $\hat{p}$ predicted in \eqref{eq9}, the concatenation of $\bm{s}_t$ and $\bm{c}_t$ is used as a query to the $\hat{p}$th row of topic memory and update $\bm{s}_t$ with the weighted sum of topic embedding in $\bm{C}$:
\begin{align}
score(t,j)&=\frac{\exp(\left[\bm{s}_t;\bm{c}_t\right]\bm{W}^t\bm{C}_{\hat{p},j})}{\sum_{j=1}^K \exp(\left[\bm{s}_t;\bm{c}_t\right]\bm{W}^t\bm{C}_{\hat{p},j})} \ 1\le j\le K 
\end{align}
and $f(\bm{s}_t)$ in \eqref{eqd} is realized by:
\begin{align}
f(\bm{s}_t) &= \bm{s}_t + \bm{V}\sum_{j=1}^K score(t,j)\bm{C}_{\hat{p},j}
\end{align}
where $\bm{W}^t \in \mathbb{R}^{2d\times d}$ and $\bm{V} \in \mathbb{R}^{d\times d}$ serves for linear projection. Futhermore, memory contexts are initialized by the pretrained word representation, but during the generating process, it should be dynamicly updated with the produced sequence to keep
recording new information, thus the topic memory can provide better guidance for the generator. We achieve this goal by computing a weight vector with a gated mechanism to weight in what degree the topic memory should be updated, then we obtain candidate state based on $\bm{s}_t'$ and $\bm{C}_{\hat{p},j}$, where $\bm{W}^u, \bm{W}^c \in \mathbb{R}^{d \times d}$:
\begin{align}
\bm{u} &= \sigma(\bm{W}^u \left[\bm{s}_t';\bm{C}_{\hat{p},j}\right]) \\
\tilde{\bm{C}}_{\hat{p},j}&= tanh(\bm{W}^c \left[\bm{s}_t';\bm{C}_{\hat{p},j}\right]) \\
\bm{C}_{\hat{p},j} &= \bm{u} \otimes \tilde{\bm{C}}_{\hat{p},j}+(\bm{1}-\bm{u}) \otimes \bm{C}_{\hat{p},j}
\end{align}

\subsection{Commonsense Enforcement}
We argue it's beneficial to make our network leverage context-related concepts. We thus implement commonsense enforcement in two aspects:
word knowledge pretraining and commonsense aware generator.\\
\textbf{Word Embedding Pretraining For Commonsense Enforcement}: We directly enrich information of our generator by pretraining word-level representation in an external commonesense KB. Note that word embedding pretraining is an off line step and is based on Graph Attention Network (GAT) ~\cite{velickovic2017graph}. For detail, see the Appendix.

\noindent \textbf{Commonsense Aware Generator}: In decoding phase, we merge neighbour nodes information in commonsense KB of generated words in the previous step to inject commonsense knowledge into our generator. For example, if "original cost" has been generated, we hope next sequence is "of the stock", other than "of the volume", for stock has the property "cost". Assume the decoder generates word $w_{t-1}$ in step $t-1$, we extract a sub-graph within two-hop paths starting from $w_{t-1}$ by Breadth First Search (BFS), as is shown in Figure \ref{fig3}. Let $\bm{e}_{ij}$ denote the path representation from node $i$ to node $j$ if $i$ and $j$ are directly connected:
\begin{figure}[htbp]
\centerline{\includegraphics[width=6cm]{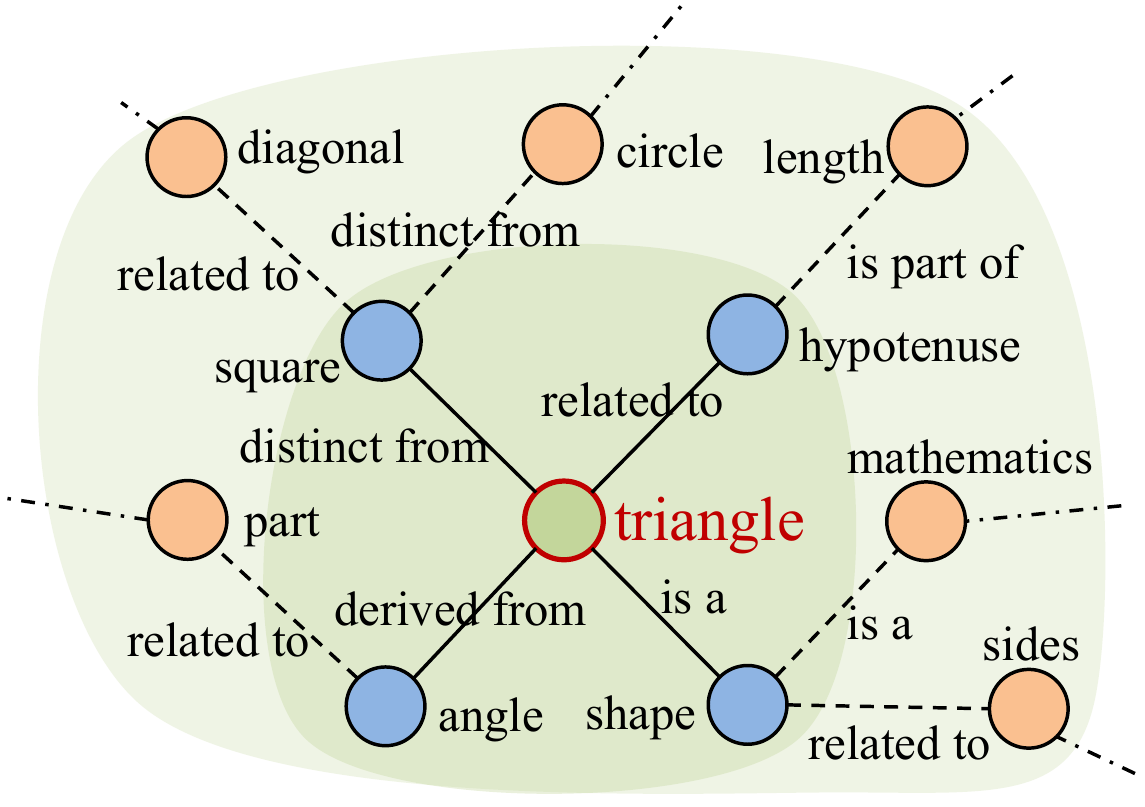}}
\caption{ Illustration of searching adjacent nodes. For word ``triangle", first-order neighbors in knowledge graph are colored in blue while second-order neighbors are colored in orange.}
\label{fig3}
\end{figure}
\vspace{-0.5cm}

\begin{align}
\bm{e}_{ij}=\phi(\bm{W}^g \left[\bm{e}_i;\bm{e}_j\right])
\end{align}
where $\bm{W}^g \in \mathbb{R}^{2d\times d}$. If $i$ and $j$ are connected via intermediate node $k$, we aggregate 
the shortest path representation from $i$ to $j$ to obtain $\bm{e}_{ij}$:
\begin{align}\label{eq21}
\bm{e}_{ij}= \alpha \phi(\bm{W}^g \left[\bm{e}_i;\bm{e}_j\right])+(1-\alpha) \sigma(\bm{e}_{ik}\otimes (\bm{U}\bm{e}_{kj}))
\end{align}
where $\bm{U}\in \mathbb{R}^{d\times d}$, $\phi(\cdot)$ is a nonlinear function, in this paper we use $tanh(\cdot)$. $\sigma(\cdot)$ is Sigmoid function. $\alpha \in \left[0,1\right]$ is a scalar to control the contribution of direct and indirect information. Denote first order neighbour set and second order neighbour set of $w_{t-1}$ as $\mathcal{N}_1(w_{t-1})$ and
$\mathcal{N}_2(w_{t-1})$, respectively. We use an attention mechanism to tend to all possible paths, i.e., we calculate the aggregate summary of $\bm{e}_{w_{t-1},j}$ when $j$ goes through $\mathcal{N}_1(w_{t-1})\cup \mathcal{N}_2(w_{t-1})$:
\begin{align}
\beta_{t-1,j} & \propto \exp(\bm{e}_{w_{t-1}}\bm{W}^b\bm{e}_{w_{t-1},j}) \\
\label{eq23} \bm{g}_{t-1}&= \sum_{j\in \mathcal{N}_1(w_{t-1})\cup \mathcal{N}_2(w_{t-1})} \beta_{t-1,j}\bm{e}_{w_{t-1},j}
\end{align}
Followed by \eqref{eq23}, to better reflect the effect of concept knowledge to word choice, we combine $\bm{e}_{w_{t-1}}$ with $\bm{g}_{t-1}$ to realize $g(\bm{e}_{w_{t-1}})$ in \eqref{eqd}:

\begin{align}
g(\bm{e}_{w_{t-1}})=GRU(\bm{e}_{w_{t-1}},\bm{H}\bm{g}_{t-1})
\end{align}

\subsection{Training Objective}
We aggregate 1): VAE loss mentioned in
\eqref{eq10} 2) auxiliary topic prediction loss $\mathcal{L}_{topic}=\mathbb{E}_{z\sim \mathcal{N}(\bm{\mu}',\bm{\sigma}'^2 \textbf{I})} p(\hat{p}|z,\bm{x})$ to obtain total loss:
\begin{align}\label{eq26}
\mathcal{L}_{total}=\mathcal{L}_{VAE}+\mu\mathcal{L}_{topic}
\end{align}
where $\mu$ is a hyperparameter.

\section{Experiments}
\subsection{Datasets}
\par Dolphin-18K ~\cite{huang-etal-2016-well} is the largest MWP dataset with various types of MWP text, while only a part of it (3154) are released. We then reuse the python script provided by ~\cite{huang-etal-2016-well} to crawl and collect data from Yahoo, which extends Dolphin-18K to 9643 samples in total. Statistic information of our data is listed in Table \ref{dataset}. We conduct some data preprocessing by deleting those equation-problem text pairs whose problem text length is longer than 45 tokens, besides, we replace those words appearing less than 2 times to $\langle$UNK$\rangle$. 

\newcommand{\tabincell}[2]{\begin{tabular}{@{}#1@{}}#2\end{tabular}}
\subsection{Motivation of Creating New Dataset}
MWP solving datasets currently used include Alg514 \citep{kushman-etal-2014-learning}, Dolphin1878 \citep{shi-etal-2015-automatically}, DRAW-1K \citep{upadhyay-chang-2017-annotating}, Dolphin18K \citep{huang-etal-2016-well}. Table \ref{dataset} gives the statistic of these datasets. 
Alg514, Dolphin1878, DRAW-1K are all public available, while neural generation models for generative tasks are usually data-hungry thus equation-MWP pairs in those datasets are insufficient. Though Dolphin18K is a large scale dataset, only a part of it (3154) are released. Moreover, existing datasets only include a certain type of MWP text, e.g., MWP text for linear equations, which restricts their practical application. We then reuse the python script provided by \citep{huang-etal-2016-well} and acquire 14943 equation-MWP text pairs in total from Yahoo !. Generally, the public available datasets can be treated as the subset of our dataset. Next, we conduct data preprocess as follows, which is beneficial to train the generation model:
\begin{itemize}
    \item We normalize the equations by replacing all the equation variables in each sample to $x,y,z$,... in order, e.g., $u+v+r=100,u-r=10$ is replaced to $x+y+z=100, \ x-z=10$.
    \item We manually correct the wrong spelling words in MWP text.
\end{itemize}

\begin{table}[h]
\small
\centering
\scalebox{0.75}{
\begin{tabular}{c c c c c}
\toprule
\textbf{Dataset} & \textbf{Size}  & \textbf{Problem Type} & \textbf{Avg EL} & \textbf{Avg Ops} \\
\midrule
Alg514 & 514 & algebra, linear & 9.67 & 5.69\\
Dolphin1878 & 1878 & number word problems & 8.18 & 4.97\\
DRAW-1K & 1000 & algebra, linear, one-variable & 9.99 & 5.85\\
Dolphin18K &  $18460^{*}$  & algebra, linear,  multi-variable  & 9.19 & 4.96  \\
\midrule
Our Dataset & 14943 & \tabincell{c}{algebra, linear/nonlinear, \\multi-variable} & 16.64  & 6.41
\\
\bottomrule
\end{tabular}}
\caption{Statistics of several existing MWP solving datasets. Avg EL, Avg Ops refer to average equation length and average numbers of operators in equations, respectively. $*$ indicates only 3154 equation-MWP pairs of Dolphin18K are available.}
\label{dataset}
\end{table}

\subsection{Model Settings}
\par The batch size for training is 32. We employ ConceptNet5 \footnote{https://github.com/commonsense/conceptnet5} to construct KB, it has 34 types of relationship in total. 2 layer graph attention network is implemented for word knowledge pretraining step. The embedding size and all hidden state size of GRU are set to 256. In problem encoder three different convolutional kernels are used and their kernel sizes are 2,3,4, respectively. To be fair, we use 1-layer GRU for both our model and baseline. For LDA we divide all samples into 9 topic types and their amount and representative words are reported in Table \ref{topic}. Each problem is associated with a topic distribution over 9 topics. The topic with the highest probability is adopted as the golden category. Meanwhile each topic contains several words and we choose top 30 words to construct topic memory for each topic. $\mu$ in (20) is set to 0.5. Weight coefficient in (16) is set to $\alpha=0.7$. We use Adam optimizer ~\cite{kingma2014adam:} to train our model, the learning rate is set to 0.0005.

\begin{table}[htbp]
\small
\centering
\begin{tabular}{c|c|c}
\toprule
\textbf{Topic type} & \textbf{Train}  & \textbf{Representative words} \\
\midrule
1 & 810 & do,people,divided,men,mean... \\
\midrule
2 & 758 & length,width,rectangle,area,inches... \\
\midrule
3 & 1573 & probability,quarter,dimes,coins,marbles...\\
\midrule
4 &  557  & sum,difference,larger,smaller,less...    \\
\midrule
5 & 1087 & solution,gallon,mixture,grams,water... \\
\midrule
6 & 633 &  interest,year,invested,dollars,rate... \\
\midrule
7 & 663 &  angles,degrees,percent,digit,increased... \\
\midrule
8 & 879 & sold,ticket,prices,children,adult... \\
\midrule
9 & 754 & speed,minutes,travels,took,plane... \\
\bottomrule
\end{tabular}
\caption{Topic classes statistics and representative words sampled from each topic}
\label{topic}
\end{table}

\vspace{-0.2cm}
\subsection{Automatic Evaluation}
\par We report automatic evaluation in five aspects: BLEU (up to bigrams) ~\cite{papineni2002bleu:}, ROUGE-L ~\cite{lin2004rouge:}, Dist-1, Dist-2, which indicates the proportion of different unigrams (bigrams) in all unigrams (bigrams), Number recall, which is used to measure how many numbers in problem text are correctly copied. Results are reported in Table~\ref{auto1}. In Table~\ref{auto1} we also present results of ablation study. We can observe 1) our model yields higher performance in all metrics compared with baselines, especially in Dist-1 and Dist-2, which proves our model can generate more diversity 
math word problems. We 
consider this is because baseline models have no guidance in topic words and knowledge, thus they tend to generate the simplest question type like ``one number is twice the second number ...". 2) taking out topic control or commonsense enhancement will both decrease evaluation scores, which verifies their effectiveness. For example, removing commonsense enhancement declines BLEU score by 24.4\%, while removing VAE \& topic memory declines BLEU score by 35.5\%.
\par We also separately compare MAGENT with our model including the same keywords as an extra input in Table~\ref{auto2}, which demonstrates our model can 
still achieve performance gain with the same input.
\vspace{-0.5cm}

\begin{table}[htbp]
\small
\centering
\caption{Statistic of datasets.}
\scalebox{0.8}{\begin{tabular}{c c c c}
\hline
 & \textbf{Train} & \textbf{Dev} & \textbf{Test} \\
\hline
\textbf{Size}& 7714 & 964 & 965 \\
\textbf{Equation Length (average)}& 16.69 & 16.23 & 16.63 \\
\textbf{Problem Length (average)} & 28.90 & 29.64 & 28.74\\
\textbf{Tokens} & 7445 & 3065 & 2875 \\
\bottomrule
\end{tabular}}
\label{dataset}
\end{table}
 \vspace{-1cm}

\begin{table}[htbp]\label{t2}

\small
\centering
\caption{ Automatic results in test dataset with BLEU, ROUGE-L (ROU), Dist-1 (D1), Dist-2 (D2) and Number Recall (NR). TP, TM and V denote the equation template, topic memory and VAE, respectively. CE includes both the pretraining step and the commonsense enforcement for the decoder. }
\scalebox{0.8}{\begin{tabular}{p{56pt} p{40pt} p{40pt} p{40pt} p{40pt} p{40pt}}
\hline

 \textbf{Model} & \textbf{BLEU} & \textbf{ROU} & \textbf{D1}($\%$) & \textbf{D2}($\%$) & \textbf{NR}($\%$) \\
\hline
Seq2seq & 0.0259 &0.2025 & 14.56& 34.99 & 47.60 \\

SeqGAN & 0.0262 & 0.1922 & 12.96 & 30.02 & 44.00 \\
DeepGCN & 0.0304 & 0.2094 & 16.81 & 45.17 & 49.21 \\
Transformer & 0.0277 & 0.2036 & 16.69 & 37.57 & 50.89 \\
\hline
\tabincell{l}{Our model } & \textbf{0.0433} & \textbf{0.2415} & \textbf{20.84} & 53.81 & 55.14 \\
w/o TP & 0.0385 & 0.2377 & 18.88 & \textbf{57.41} & \textbf{55.84} \\
\tabincell{l}{w/o CE} &0.0327 & 0.2273 & 18.75 & 51.19 &54.42 \\
 w/o TM & 0.0345 & 0.2256 & 20.00 & 55.00 & 54.31 \\
 w/o V \& TM & 0.0280 & 0.2141 & 18.79 & 50.05 & 53.82 \\
\hline

\end{tabular}}
\label{auto1}
\end{table}
 
\begin{table}[htbp]
\small
\centering
\caption{Comparison between our model with keywords (KW) and MAGNET in automatic results}
\renewcommand\tabcolsep{4.0pt}
\scalebox{0.8}{\begin{tabular}{p{83pt} p{40pt} p{40pt} p{40pt} p{40pt} p{40pt}}
\hline

 \textbf{Model} & \textbf{BLEU} & \textbf{ROU} & \textbf{D1}($\%$) & \textbf{D2}($\%$) & \textbf{NR}($\%$) \\
\hline
MAGNET & 0.0976 &0.3793 & \textbf{21.72} & 57.22 & 42.62 \\
\hline
Our model (KW) & \textbf{0.1152} & \textbf{0.4006} & 18.81 & \textbf{58.85} & \textbf{51.50} \\
\hline
\end{tabular}}
\label{auto2}
\end{table}

\vspace{-1cm}

 \subsection{Human Evaluation}
 \par Automatic metrics such as BLEU and ROUGE only focus on $n$-gram similarity, but fail to measure true generation quality (i.e., if topic drifting occurs). We invite three human annotators to judge generation quality in four aspects. 1) \textbf{Fluency (Flu)}: it mainly judges whether the problem text is fluent, i.e., whether the generated problem text has some grammar errors. 2) \textbf{Coherence (Coh)}: it weights if the problem text is coherent in text-level; 3) \textbf{Solvability-1 (S1)}: as our target is a math word problem, we should pay attention to whether the problem text can be solved, i.e., in what percentage we can set up the same (or equivalent) equations and solve them according to the generated problem text; 4) \textbf{Solvability-2 (S2)} is a more relaxed criterion compared with Solvability-1, it only requires the text produced is a valid math problem and could be solved regardless what equations could be set. We randomly select 50 generated MWP texts and score them in five grades. The scores are projected to 1$\sim$5, where higher score implies better performance (for solvability we use percentage). We report the average scores in Table~\ref{table:human-evaluation}.
 
 \par Table~\ref{table:human-evaluation} (\emph{upper}) confirms our proposed model receives significant higher score in coherence and solvability, we assume this is because our model restricts the problem text into a certain topic and provides related words for reference. 

\par In Table~\ref{table:human-evaluation} (\emph{bottom}) we report comparison between our model with keywords and MAGNET. Human scores reflect that our method achieves 12\% relative improvement over MAGNET in Solvability-1. Especially, with keywords fed into the model, the problem of topic drifting is no longer notable for both our model and MAGNET.

\begin{table}[h]
\caption{Human evaluation results: comparison between the proposed model and baseline models. }
\setlength{\abovecaptionskip}{-5pt}
\setlength{\belowcaptionskip}{-10pt}
\small
\centering
\scalebox{0.8}{\begin{tabular}{c p{26pt} p{26pt} p{36pt} p{36pt} }
\hline
 & \textbf{Flu} & \textbf{Coh} &\textbf{S1}(\%) & \textbf{S2}(\%) \\
\hline
Our model & \textbf{4.03} & \textbf{4.02} & \textbf{35} & \textbf{55} \\
\hline
Seq2seq & 3.78 & 3.48 & 23 & 34 \\
SeqGAN & 3.75 & 3.28 & 20 & 40 \\
 
DeepGCN & 3.61 & 3.55 & 29 & 52 \\
Transformer & 3.80 & 3.53 & 20 &45 \\
\hline
\hline
MAGNET & 4.00 & 4.33 & 44 & \textbf{76} \\
\hline
Our model (KW) & \textbf{4.27} & \textbf{4.60} & \textbf{56} & 74 \\
\hline \\
\end{tabular}}
\label{table:human-evaluation}
\end{table}

\subsection{Case Study}
\par Table~\ref{table:case-study} shows some math word problems generated by different models. It's easy to show problem text generated by Seq2seq suffers from lack of coherence, e.g., in the above case, the baseline result talks about different topics in the same sentence. As a comparison, our generator discusses the same topic and generates words around this topic. What's more, the topic of problem text generated by our proposed model is highly consistent with reference answer, which verifies the effectiveness of the proposed model.
\par We can also observe commonsense violation appears in baseline results, for example, "chemist has a perimeter" and "geometric is 4 more than" are obviously illogical. Relatively speaking, MWP text generated by our model, is more in line with commonsense, such as "the hypotenuse of a right triangle". These results reflect that our model can benefit from both the topic consistency maintaining and commonsense enforcement mechanism.

\begin{table*}[ht]
\caption{Three examples of math word problems generated by different models. Transformer is abbreviate to Trans. Topic words in the left column indicate the overlap between selected topic words and the generated MWP text, which is also highlighted in the right column. CG reflects the reasoning procedure adopted by the decoder. }
\small
\centering
\scalebox{0.65}{
\begin{tabular}{|p{9.3cm} |p{9.4cm}|}
\toprule
 \tabincell{l}{\textbf{Equation}: $equ : 4 * ( x - y ) = 800 \ equ : 2 * ( x + y ) = 800$ \\ \textbf{Reference}: An airplane travels 800 miles against the wind in 4 \\ hrs and makes the return trip with the same wind in 2 hrs . Find \\the speed of the wind. \\ \textbf{Topic Words}: travels, miles, speed \\ \textbf{CG}: travel$\xrightarrow[]{RelatedTo}$ trip travel $\xrightarrow[]{RelatedTo}$ take}& 
 \tabincell{l}{\textbf{Ours}: A plane \colorbox{pink}{travels} 800 \colorbox{pink}{miles} in 4 hours . the return trip\\ against the same wind took 4.5 hours. Find the \colorbox{pink}{speed} of the\\ current wind. \\
 \textbf{Seq2seq}: A chemist has a perimeter of 80 cm, the area of the \\rectangle is 800 m. what is the length of the rectangle.\\ \textbf{SeqGAN}: The perimeter of a rectangle is 800 inches. Find the \\length.\\
 \textbf{DeepGCN}: A man has 800 more than four times as old as his \\son . If the current is 800 m. Find the speed of the plane. \\ \textbf{Trans}: The sum of two numbers is 800. The sum of their \\squares is 25\% . Find the numbers.}
\\
\midrule
\tabincell{l}{\textbf{Equation}: $ equ : 0.1 * x + 0.05 * y = 1.95 \ equ : y = x + 3
$ \\ \textbf{Reference}: Ken has \$ 1.95 in nickels and dimes . There are 3 \\more nickels than dimes. How many of each does he have? \\ \textbf{Topic Words}: nickels, dims, ticket  \\ \textbf{CG}: nickel$\xrightarrow[]{IsA}$coin$\xrightarrow[]{RelatedTo}$dime }& 
 \tabincell{l}{\textbf{Ours}: Arnold has some \colorbox{pink}{nickels} and \colorbox{pink}{dimes}. If he made a \\number of \$ 3 .50 . If she had 3 more nickels than dimes. Find\\ the amount of each \colorbox{pink}{ticket}.\\
 \textbf{Seq2seq}: A total of 1.95 seats is in the ratio of 10\%. If the \\total area of the coins is 1.95. Find\\ \textbf{SeqGAN}: A carpet is 3 times as many more than the other.\\ The total value is 3.\\
 \textbf{DeepGCN}: Dan invested \$ 1.95 . part a t 10\% annual interest. \\the total value of the cost of a 10\% salt solution is added to  \\ \textbf{Trans}: A computer has a mixture of \$ 3 . 000 and a mixture\\ contains 5\% profit of 5\% . How much does each have.}
 \\
\midrule
\tabincell{l}{\textbf{Equation}: $ equ : 0.1 * x + 0.05 * y = 1.95 \ equ : y = x + 3
$ \\ \textbf{Reference}: A car radiator contains 10 liters of 30\% antifreeze\\ solution. How many liters will have to be replaced with the pure \\antifreeze if the resulting solution is to be 50\% antifreeze? \\ \textbf{Topic Words}: acid, solution   \\ \textbf{CG}: acid $\xrightarrow[]{IsA}$liquid $\xrightarrow[]{RelatedTo}$ solution}& 
 \tabincell{l}{\textbf{Ours}: How many liters of a 30\% \colorbox{pink}{acid solution} must be added \\to 10 liters of 20\% solution on the mixture to make l0 litters of \\a 50\% solution. \\
 \textbf{Seq2seq}: A raditor contains 50\% nitric acid. If we take the same\\ place to be 10. What is the total number of water.\\
 \textbf{SeqGAN}: A boat travels 10 miles per hour. How much would a \\20\% acid solution to be worth in 100 account \\ 
 \textbf{DeepGCN}: A $<$UNK$>$ radiator is \$ 10 . 00 a t a certain number.\\ and the rest a t the same time .  \\ \textbf{Trans}: A car traveled in a car point is going a t a constant speed. \\If the car going a car point in the car . about 0 mph . how many \\miles apart.}
 \\
\bottomrule
\end{tabular}}
\label{table:case-study}
\end{table*}

\section{Conclusion}

\par We propose a novel model and a dataset for generating MWP from equations. Our model can effectively encode different types of math tokens in equations and reduce the gap between abstract math tokens and generated natural language text. It is also very useful in tackling the topic drifting and commonsense violation problems. Experiments on our dataset show our model significantly outperforms baseline models.

\end{document}